\documentclass[10pt,twocolumn,letterpaper]{article}

\usepackage{cvpr}              

\definecolor{cvprblue}{rgb}{0.21,0.49,0.74}
\usepackage[pagebackref,breaklinks,colorlinks,allcolors=cvprblue]{hyperref}
\usepackage{multirow}
\usepackage{verbatim}
\usepackage{comment}
\usepackage{adjustbox}
\usepackage{wrapfig}

\usepackage{graphicx}
\usepackage{amsmath}
\usepackage{amssymb}
\usepackage{booktabs}
\usepackage{bm}

\usepackage{url}            
\usepackage{amsfonts}       
\usepackage{nicefrac}       
\usepackage{microtype}      
\usepackage{xcolor}         

\usepackage{array}
\usepackage{xspace}

\usepackage[accsupp]{axessibility}

\newcolumntype{L}[1]{>{\raggedright\let\newline\\\arraybackslash\hspace{0pt}}m{#1}}
\newcolumntype{C}[1]{>{\centering\let\newline\\\arraybackslash\hspace{0pt}}m{#1}}
\newcolumntype{R}[1]{>{\raggedleft\let\newline\\\arraybackslash\hspace{0pt}}m{#1}}

\newcommand{\ignore}[1]{}

\makeatletter
\DeclareRobustCommand\onedot{\futurelet\@let@token\@onedot}
\def\@onedot{\ifx\@let@token.\else.\null\fi\xspace}

\makeatother

\definecolor{MyDarkBlue}{rgb}{0,0.08,1}
\definecolor{MyDarkGreen}{rgb}{0.02,0.6,0.02}
\definecolor{MyDarkRed}{rgb}{0.8,0.02,0.02}
\definecolor{MyDarkOrange}{rgb}{0.40,0.2,0.02}
\definecolor{MyPurple}{RGB}{111,0,255}
\definecolor{MyRed}{rgb}{1.0,0.0,0.0}
\definecolor{MyGold}{rgb}{0.75,0.6,0.12}
\definecolor{MyDarkgray}{rgb}{0.66, 0.66, 0.66}

\definecolor{turquoise}{cmyk}{0.65,0,0.1,0.3}

\newcommand{\model}{MVLift\xspace}

\newcommand{\myparagraph}[1]{\vspace{-10pt}\paragraph{#1}}

\def\paperID{4186} 
\def\confName{CVPR}
\def\confYear{2025}

\title{Lifting Motion to the 3D World via 2D Diffusion}

\author{Jiaman Li \qquad C. Karen Liu\footnotemark[2] \qquad Jiajun Wu\footnotemark[2] \\
Stanford University\\
{\tt\small \{jiamanli,karenliu,jiajunwu\}@cs.stanford.edu}
}

\begin{document}
\twocolumn[{%
\renewcommand\twocolumn[1][]{#1}%
\maketitle
\vspace{-10mm}
\begin{center}
    \centering
    \includegraphics[width=\textwidth]{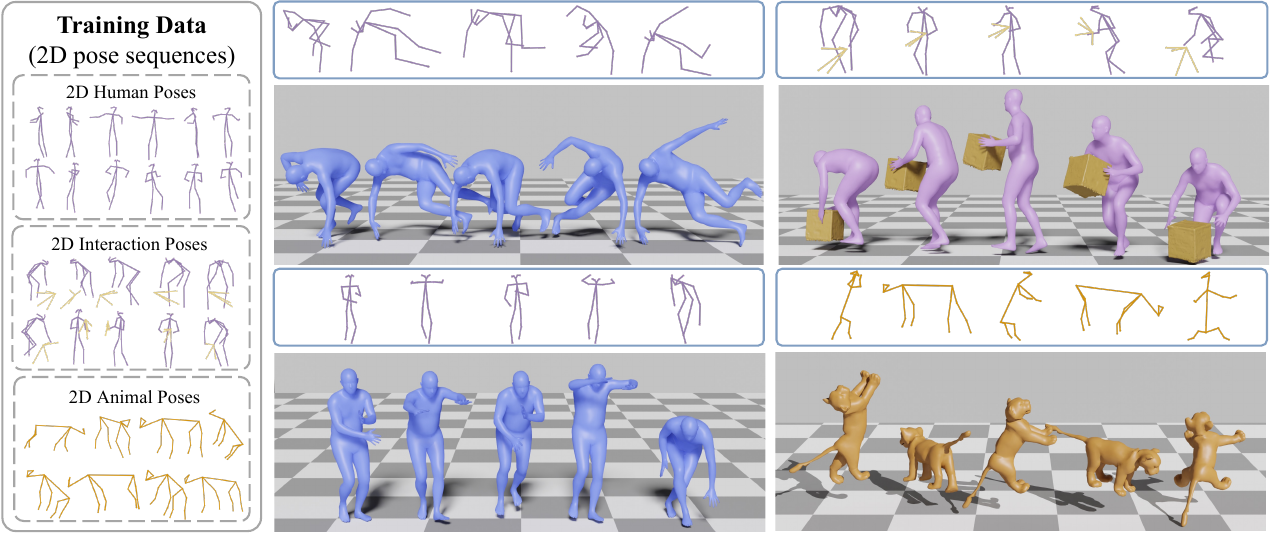}
    \vspace{-6mm}
    \captionof{figure}{Our framework, \model, can be trained only on 2D pose sequences and generate 3D motions including joint rotations and root trajectories in the world coordinate system. The approach generalizes to the various domains of human poses, interactions, and animal poses.}
    \label{figure1}
\end{center}%
}]

\setcounter{figure}{1}

\begin{abstract}

\vspace{-2mm}
\begin{NoHyper}
\footnotetext{\footnotemark[2] indicates equal advising.}
\end{NoHyper}
Estimating 3D motion from 2D observations is a long-standing research challenge. Prior work typically requires training on datasets containing ground truth 3D motions, limiting their applicability to activities well-represented in existing motion capture data. This dependency particularly hinders generalization to out-of-distribution scenarios or subjects where collecting 3D ground truth is challenging, such as complex athletic movements or animal motion. We introduce \model, a novel approach to predict global 3D motion---including both joint rotations and root trajectories in the world coordinate system---using only 2D pose sequences for training. Our multi-stage framework leverages 2D motion diffusion models to progressively generate consistent 2D pose sequences across multiple views, a key step in recovering accurate global 3D motion. \model generalizes across various domains, including human poses, human-object interactions, and animal poses. Despite not requiring 3D supervision, it outperforms prior work on five datasets, including those methods that require 3D supervision.

\end{abstract}



\vspace{-5mm}
\section{Introduction}
\vspace{-1mm}

Generating 3D motion is critical for applications in computer graphics, embodied AI, and robotics. High-quality 3D motion datasets~\cite{AMASS,li2023object,jiang2024scaling,araujo2023circle,GRAB:2020}, essential for developing generative models that synthesize human movement, are typically obtained using advanced motion capture systems. However, these systems are often confined to lab environments and are costly to scale, particularly for capturing complex motions that require specialized expertise, such as dancing, yoga, and gymnastics. Extensive work has studied monocular video pose estimation to predict 3D human motion from 2D observations without the need for elaborate mocap setups (\cite{kocabas2019vibe,zhu2023motionbert, shin2024wham}). Despite advancements, even state-of-the-art methods (\cite{shin2024wham}) still heavily rely on motion capture datasets and videos with paired 3D ground truth. This dependence limits their ability to generate accurate global 3D motions for out-of-distribution motions—those not well-represented in existing 3D datasets.

To overcome the challenge of generating out-of-distribution 3D motions, recent studies have explored methods to leverage in-domain 2D poses and generate 3D poses using only 2D poses during training. Some efforts train models exclusively on 2D poses of a single step, often resulting in unsmooth motions and noticeable artifacts when applied to 2D pose sequences~\cite{wandt2022elepose}. Recent work has employed a pre-trained unconditional 2D motion diffusion model to generate 3D motions from sampled noise~\cite{kapon2024mas}, demonstrating the ability to produce complex movements not typically covered by existing datasets, such as gymnastics, horse motion, and basketball. However, this approach does not account for the global translation of 3D motion. In this work, we address the challenge of predicting global 3D motion for out-of-distribution scenarios using only domain-specific 2D pose data (e.g., human, animal, or interaction). Unlike prior work that primarily focuses on generating 3D motion with a fixed root joint, our goal is to accurately predict realistic 3D motion with root trajectories in the world coordinate system from 2D pose sequence inputs.

The problem of estimating global 3D motion from 2D pose sequences can be effectively addressed using a single-stage network when 3D training data is available~\cite{zhu2023motionbert}. However, training a single-stage model using solely 2D poses is challenging due to the lack of direct supervision for either 3D motion or consistent multi-view 2D poses. To address this fundamental challenge, we propose a multi-stage approach, \model, which progressively establishes multi-view consistency through 2D motion diffusion models, enabling accurate 3D motion recovery without 3D training data.

\model consists of four stages. First, we train a line-conditioned diffusion model that learns to predict 2D pose sequences following epipolar lines. These epipolar lines, derived from the fundamental matrix relating two camera views, ensure corresponding points in different views lie along corresponding lines, enforcing basic geometric consistency.
Second, we develop a joint optimization method for multi-view 2D motion sequences using two objectives: (1) a multi-view consistency objective to ensure geometric relationships across views, and (2) Score Distillation Sampling (SDS) based on our trained line-conditioned diffusion model to maintain motion realism. While this approach generates plausible multi-view 2D sequences that enable 3D motion recovery, the optimization process cannot guarantee perfect view consistency, affecting the accuracy of reconstructed 3D motions.
To address this limitation, we leverage these optimized results to create a synthetic dataset. Although the optimized 3D motions may not perfectly align with input 2D sequences, they preserve motion realism. By reprojecting these synthetic 3D motions to different views, we obtain strictly consistent multi-view 2D sequences. Finally, we use these sequences to train an efficient diffusion model that directly generates multi-view consistent 2D sequences in a single forward pass.

Our work makes three contributions. First, we introduce a novel framework, \model, for estimating global 3D motion from a single-view 2D pose sequence without any 3D training data, addressing a fundamental limitation in existing approaches. Second, we demonstrate how multi-view consistency can be progressively established through 2D motion diffusion, providing a new perspective on 3D motion estimation. Third, we show that \model generalizes across various domains (humans, animals, and interactions) and significantly outperforms existing methods on five datasets, even those trained with 3D supervision. 

\section{Related Work}

\paragraph{Human Pose Estimation from 2D Poses.}
With the advent of datasets containing paired 2D videos and ground truth 3D motions~\cite{ionescu2013human3,sigal2010humaneva,mehta2017monocular}, various approaches have been proposed to learn the mapping between 2D human joints and 3D human joints, including fully connected networks~\cite{martinez2017simple}, temporal convolution networks~\cite{pavllo20193d,pavllo20193d}, GCN-based networks~\cite{wang2020motion,cai2019exploiting,ci2019optimizing}, and more recently, Transformer-based models~\cite{li2022mhformer,zhang2022mixste,zheng20213d,shan2022p}. The typical paradigm involves training neural networks to regress 3D poses from 2D pose inputs. However, this is constrained by the limited availability of paired data and is often incapable of predicting accurate 3D poses outside the distribution of paired data. To address this problem, a recent work, MotionBERT~\cite{zhu2023motionbert}, leverages the large-scale mocap dataset~\cite{AMASS} to generate paired 2D and 3D data for training, showcasing superior results. Nevertheless, the issue of limited data availability persists. To overcome this data challenge, some work explores the problem of estimating 3D poses by training solely on in-domain 2D poses, without any 3D ground truth poses. For example, ElePose~\cite{wandt2022elepose} predicts 3D pose and camera pose simultaneously with a reprojection loss and a pose prior loss. MAS~\cite{kapon2024mas} optimizes 3D motion using the learned 2D motion diffusion model. However, these methods do not account for the global trajectory of 3D motion. In our work, we use a 2D motion sequence as input and predict global 3D motion, including joint rotations and a root trajectory in the world coordinate system, without training on any 3D motion data. 

\myparagraph{Human Pose Estimation from Videos.}
Estimating SMPL~\cite{smpl} parameters from images~\cite{smplify,kanazawa2018end,kolotouros2019learning,smplx,kocabas2021pare,kocabas2021spec,li2021hybrik,li2023niki,zhang2023pymaf} or videos~\cite{kanazawa2019learning,kocabas2019vibe,wan2021encoder,shen2023global,wei2022capturing,shin2024wham} has been extensively studied in recent years. Reconstructing human mesh from images has been improved by incorporating optimization techniques~\cite{smplify} into the learning pipeline~\cite{kolotouros2019learning} and leveraging more powerful backbones~\cite{goel2023humans} such as Transformers~\cite{vaswani2017attention}. In terms of recovering human mesh from videos, prior works~\cite{kocabas2019vibe, luo20203d, choi2021beyond} employ recurrent neural networks to integrate temporal information and produce smooth 3D motion predictions. However, these works do not predict global root trajectories. Recently, there has been growing attention on predicting global 3D motion from videos~\cite{zhang2023real,yuan2022glamr, ye2023decoupling,sun2023trace,kocabas2024pace,shin2024wham}. Some works leverage the success of local 3D pose estimation as initialization and use motion priors to further optimize global pose and camera parameter predictions~\cite{yuan2022glamr, ye2023decoupling}. A recent work, WHAM~\cite{shin2024wham}, developed a pipeline that generates 3D motion in the world coordinate frame with a single forward pass. However, the global trajectory prediction module in this pipeline was trained on datasets containing ground truth 3D motion~\cite{AMASS,ionescu2013human3,mehta2017monocular}, which constrains its ability to predict accurate root trajectories for motions outside of the data distribution. In this work, we generate global 3D motions from 2D pose sequences without RGB data input. Our approach does not rely on any 3D motion data or paired video and 3D data for training, leveraging only 2D pose sequences extracted from monocular videos. The approach generalizes to various domains such as animal motions and human-object interactions, lifting 2D pose sequences to global 3D motion.

\myparagraph{Multi-View Image Generation.}
Based on a large-scale 3D object dataset~\cite{deitke2023objaverse}, considerable research has explored generating 3D objects through novel view image synthesis. Zero-1-to-3~\cite{liu2023zero} pioneered the approach of using a single image and a specified camera pose as inputs to predict the corresponding image from a new camera view. Subsequent works~\cite{chan2023generative,shi2023mvdream,liu2023syncdreamer,long2024wonder3d,shi2023zero123++,xu2023dmv3d,tang2023make,tewari2023diffusion} have advanced the field by directly training multi-view image diffusion models that can simultaneously generate consistent multi-view images. Recent efforts have also extended this idea to the more challenging task of multi-view video generation~\cite{kuang2024collaborative,van2024generative}. Inspired by these advancements, we reformulate the prediction of 3D motion as the generation of multi-view 2D motion sequences. Unlike prior efforts, which require synchronized multi-view images during training, our work circumvents the need for consistent multi-view 2D motion sequences. Instead, we train exclusively on single-view 2D motion sequences extracted from monocular videos.

\section{\model}
Our goal is to estimate global 3D motion, including both joint rotations and root trajectories in the world coordinate system, from a single-view 2D pose sequence. For motions well-represented in existing 3D datasets, this task can be addressed using a single-stage network trained with 3D supervision, as demonstrated in prior work~\cite{zhu2023motionbert}. However, for many out-of-distribution scenarios such as complex athletic movements or animal motions, 3D motion data and multi-view recordings are often unavailable or difficult to collect. In this case, a single-stage approach becomes infeasible due to the lack of direct 3D supervision. To address this challenge, we propose a multi-stage framework, \model, which only requires readily available single-view 2D data and leverages learned 2D motion diffusion priors to progressively establish multi-view consistency for 2D pose sequences across different views, enabling accurate reconstruction of realistic 3D motions without requiring any 3D supervision.

The key insight is that while a single 2D sequence provides limited 3D information, a diffusion model trained on diverse 2D motions can learn rich priors about how poses appear from different views. We use the learned 2D diffusion priors combined with geometric constraints to progressively strengthen the consistency of multi-view 2D sequences.

\begin{figure*}[t!]
\begin{center}
\vspace{-5mm}
\includegraphics[width=\textwidth]{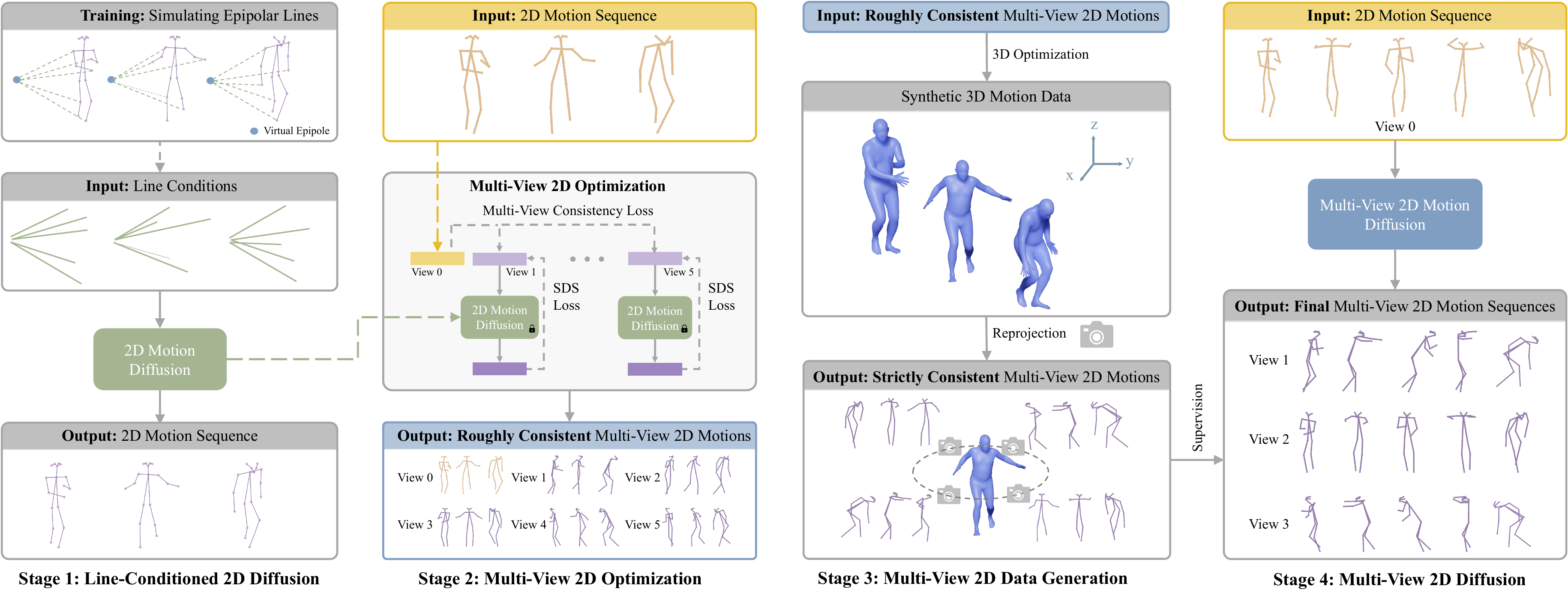}
\end{center}
\vspace{-6mm}
\caption{Overview of our multi-stage framework. In Stage 1, we train a 2D motion diffusion model conditioned on simulated epipolar lines. Stage 2 utilizes this model to optimize multi-view 2D motion sequences, achieving only roughly consistent sequences. These sequences are used for 3D motion optimization in Stage 3, and the synthetic 3D data is then reprojected into strictly consistent multi-view 2D sequences. In Stage 4, we train a multi-view 2D motion diffusion model on these data to efficiently generate consistent 2D sequences across views.}
\label{fig:overview}
\vspace{-5mm}
\end{figure*}

As illustrated in Figure~\ref{fig:overview}, our framework operates in four stages, each strengthening multi-view consistency in different ways. In Stage 1, we train a line-conditioned diffusion model that generates 2D pose sequences respecting epipolar line constraints. While this ensures basic pairwise consistency between views, achieving global multi-view consistency requires jointly considering all views. In Stage 2, we establish stronger multi-view consistency through an optimization method that directly optimizes multi-view 2D sequences using explicit multi-view consistency objectives, while maintaining motion realism through Score Distillation Sampling (SDS) based on our trained diffusion model. Although this optimization approach generates plausible multi-view 2D sequences, achieving perfect geometric consistency across all views through optimization alone remains challenging. To overcome this limitation, we leverage these roughly consistent multi-view sequences in Stage 3 to recover plausible 3D motions through optimization with 2D reprojection objectives. By reprojecting these recovered 3D motions to multiple views, we obtain a synthetic dataset of strictly consistent multi-view 2D sequences. Finally, in Stage 4, we train a specialized diffusion model to directly generate multi-view consistent 2D sequences in a single forward pass.

This progression from pairwise geometric constraints to global optimization to learned multi-view generation allows us to obtain high-quality 3D motion estimation without any ground truth 3D data for training. The following sections detail each component of our framework and how they work together to achieve 3D motion reconstruction.



\subsection{Line-Conditioned 2D Motion Diffusion Model}
We denote a 2D motion sequence as $\bm{X} \in \mathbb{R}^{T \times J \times 2}$, where $T$ is the number of time steps and $J$ is the number of joints. Each joint position at time step $t$ is denoted by $(x_{t}^{j}, y_{t}^{j})$. We define line conditions $\bm{L} \in \mathbb{R}^{T \times J \times 3}$, where each joint's line is parameterized by coefficients $(a_{t}^{j}, b_{t}^{j}, c_{t}^{j})$ following the equation $a_{t}^{j}x_{t}^{j} + b_{t}^{j}y_{t}^{j} + c_{t}^{j} = 0$. Given line conditions, we learn to generate a 2D pose sequence where each predicted joint position $(\hat{x}_{t}^{j}, \hat{y}_{t}^{j})$ aligns with its input line.


\myparagraph{Epipolar Line Generation.}
Epipolar geometry provides geometric constraints that help ensure multi-view consistency in our approach. During testing, given a 2D pose sequence $\mathbf{X}$ and a specified (rather than real) camera intrinsics matrix $\mathbf{K}$, we compute epipolar lines across different camera views using essential matrices $\mathbf{E}_v \in \mathbb{R}^{3 \times 3}$, where $v \in {1, ..., 5}$ indexes the camera views. These essential matrices encode the relative rotations and translations between views and are used to compute fundamental matrices as $\mathbf{F}_v = \mathbf{K}^{-T} \mathbf{E}_v \mathbf{K}^{-1}$. For a 2D point $\mathbf{x} = (x, y, 1)^T$ in the input view, its corresponding epipolar line in view $v$ is computed as $\mathbf{l}_v = \mathbf{F}_v \mathbf{x}$, with coefficients $(a_v, b_v, c_v)$ defining the line equation $a_vx + b_vy + c_v = 0$. During training, since only single-view 2D pose sequences are available, we simulate epipolar constraints by randomly sampling a virtual epipole for each sequence. For each joint $j$ at timestep $t$, we create a line connecting it to this virtual epipole, resulting in line coefficients $(a_t^j, b_t^j, c_t^j)$ that serve as conditions for training our line-conditioned diffusion model.

\myparagraph{Conditional Diffusion Model.}
Given line conditions $\bm{L}$, we train a conditional diffusion model that learns to generate 2D motion sequences that respect epipolar geometry constraints. The diffusion model consists of a forward diffusion process and a reverse diffusion process. The forward diffusion process progressively adds noise to the data representation $\bm{X}$ until, after $n$ steps, it resembles Gaussian noise. The reverse diffusion process involves learning a neural network that denoises the data representation starting from a noisy data state $\bm{X}_{n}$. Each denoising step is represented as,
\begin{equation}
\label{eq:reverse_diffusion_step}
    p_{\theta}(\bm{X}_{n-1}|\bm{X}_{n}, \bm{L}) := \mathcal{N}(\bm{X}_{n-1}; \bm{\mu}_{\theta}(\bm{X}_n, n, \bm{L}), \bm{\Sigma}_{n}),
\end{equation}
where $\theta$ represents the parameters of a neural network, $\bm{\mu}_{\theta}$ represents the predicted mean, and $\bm{\Sigma}_{n}$ is a fixed variance. Predicting the mean is re-parameterized as predicting the clean data representation $\bm{X}_{0}$. The training loss is defined as a reconstruction loss following prior work~\cite{ho2020denoising}, 
\begin{equation}
\mathcal{L}_{} =  \mathbb{E}_{\bm{X}_0, n}||\hat{\bm{X}}_{\theta}(\bm{X}_{n}, n, \bm{L}) - \bm{X}_{0}||_{1}.    
\end{equation}
We implement the denoising network using a transformer architecture~\cite{vaswani2017attention}, which has proven effective for motion synthesis~\cite{tevet2022human,tseng2022edge,li2023object}. The line conditions $\bm{L}$ are incorporated by concatenating them with the noisy pose features $\bm{X}_{n}$ along the feature dimension before feeding into the transformer.

\myparagraph{Line Matching Loss.}
We introduce an additional loss to ensure the predicted 2D joint positions conform to the input line conditions. This loss, defined as the perpendicular distance from each joint to its associated line, encourages the generated joints to satisfy the geometric constraints imposed by these lines. The loss is formulated as
\begin{equation}
    \mathcal{L}_{\text{line}} = \sum_{t=1}^{T}\sum_{j=1}^{J} |a_{t} ^{j} \hat{x}_{t}^{j} + b_{t}^{j} \hat{y}_{t}^{j} + c_{t}^{j}|, 
\end{equation}
where $(\hat{x}_{t}^{j}, \hat{y}_{t}^{j})$ are predictions of the $j$-th joint at time step $t$. 

\subsection{Multi-View 2D Motion Optimization}
 
While our line-conditioned diffusion model can generate 2D pose sequences for unobserved views using epipolar line constraints, these generated sequences may lack consistency across all views. This occurs because epipolar geometry only enforces pairwise constraints between each unobserved view and the input view, without ensuring global multi-view consistency. Such inconsistencies can lead to suboptimal 3D reconstruction results.
To address this limitation, we propose a multi-view optimization strategy that jointly optimizes 2D motion sequences $\{\bm{X}_{k}\}, k = 1, 2, ..., 5$ across all unobserved views. The camera views are arranged in a circular configuration around the origin, with the input view and five unobserved views separated by 60-degree intervals. Given this structured camera setup, our approach combines two optimization objectives: one objective is to ensure each 2D motion sequence is realistic, achieved using Score Distillation Sampling (SDS), and the other is to maintain consistency among different camera views, achieved through our designed multi-view consistency loss. We detail each optimization objective as follows.

\myparagraph{Score Distillation Sampling.} We adapt Score Distillation Sampling (SDS) proposed in DreamFusion~\cite{poole2022dreamfusion} to optimize multi-view 2D pose sequences. The goal is to ensure that the 2D pose sequence in each view conforms to the distribution determined by the learned line-conditioned 2D motion diffusion model. The gradient for SDS is computed as follows:
\begin{equation}
    \nabla_{\Phi_{2D}} \mathcal{L}_{\text{SDS}} = \mathbb{E}_{n, \epsilon}   [\omega(n)(\epsilon_{\theta}(\bm{X}_{n}, n, \bm{L}) - \epsilon)],
\end{equation}
where the weight $\omega(n)$ is determined by the scheduler of the diffusion model, $\bm{X}_{n}$ represents a perturbed 2D pose sequence at noise level $n$, and $\epsilon$ is random noise.

\myparagraph{Multi-View Consistency Loss.}  
We introduce a multi-view consistency loss to enforce geometric constraints across different views. Given an input view and five unobserved views ($V=6$ total views), we form $M = \binom{6}{2} = 15$ pairs by considering all possible combinations of two different views. The loss is formulated as
\begin{equation}
\mathcal{L}_{\text{multi-view}} = \frac{1}{2M} \sum_{m=1}^{M} \left( \mathcal{L}_{\text{line}}^{(v \rightarrow w)} + \mathcal{L}_{\text{line}}^{(w \rightarrow v)} \right),
\end{equation}
where $\mathcal{L}_{\text{line}}^{(v \rightarrow w)}$ measures how well the joints in view $w$ satisfy the epipolar constraints derived from view $v$:
\begin{equation}
\mathcal{L}_{\text{line}}^{(v \rightarrow w)} = \sum_{t=1}^{T}\sum_{j=1}^{J} |a(v, w) x_{j,t}^w + b(v, w) y_{j,t}^w + c(v, w)|.
\end{equation}
Here, $(x_{j,t}^w, y_{j,t}^w)$ denotes the position of joint $j$ at time $t$ in view $w$, and $(a(v, w), b(v, w), c(v, w))$ are the coefficients of the epipolar line in view $w$ computed from the corresponding point in view $v$ using the fundamental matrix $\mathbf{F}_{v,w}$. By computing this loss between all pairs of views, we encourage geometric consistency across the entire multi-view setup.

\subsection{Synthetic Multi-View 2D Data Generation}
The prior stage produces roughly consistent multi-view 2D pose sequences that enable plausible 3D motion reconstruction, although they may deviate from the input 2D pose sequence. In this stage, we create a synthetic dataset of strictly consistent multi-view 2D pose sequences using these optimized results. The process consists of three steps. First, we recover 3D joint positions by minimizing reprojection errors across all optimized multi-view 2D sequences. Following~\cite{kapon2024mas}, we then fit SMPL parameters to these 3D joints using VPoser~\cite{SMPL-X:2019}, which provides plausible joint rotations and root translations. Finally, we reproject each fitted 3D motion sequence into four camera views (arranged with 90-degree intervals) to obtain strictly consistent multi-view 2D pose sequences. The resulting synthetic dataset serves as training data for our multi-view 2D motion diffusion model described in the next section.

\subsection{Multi-View 2D Motion Diffusion Model}
Based on the synthetic multi-view 2D motion dataset obtained in the prior stage, we propose a multi-view 2D motion diffusion model to generate consistent multi-view 2D motion sequences simultaneously from a single 2D sequence input. 

\begin{figure}[t!]
\begin{center}
\includegraphics[width=\linewidth]{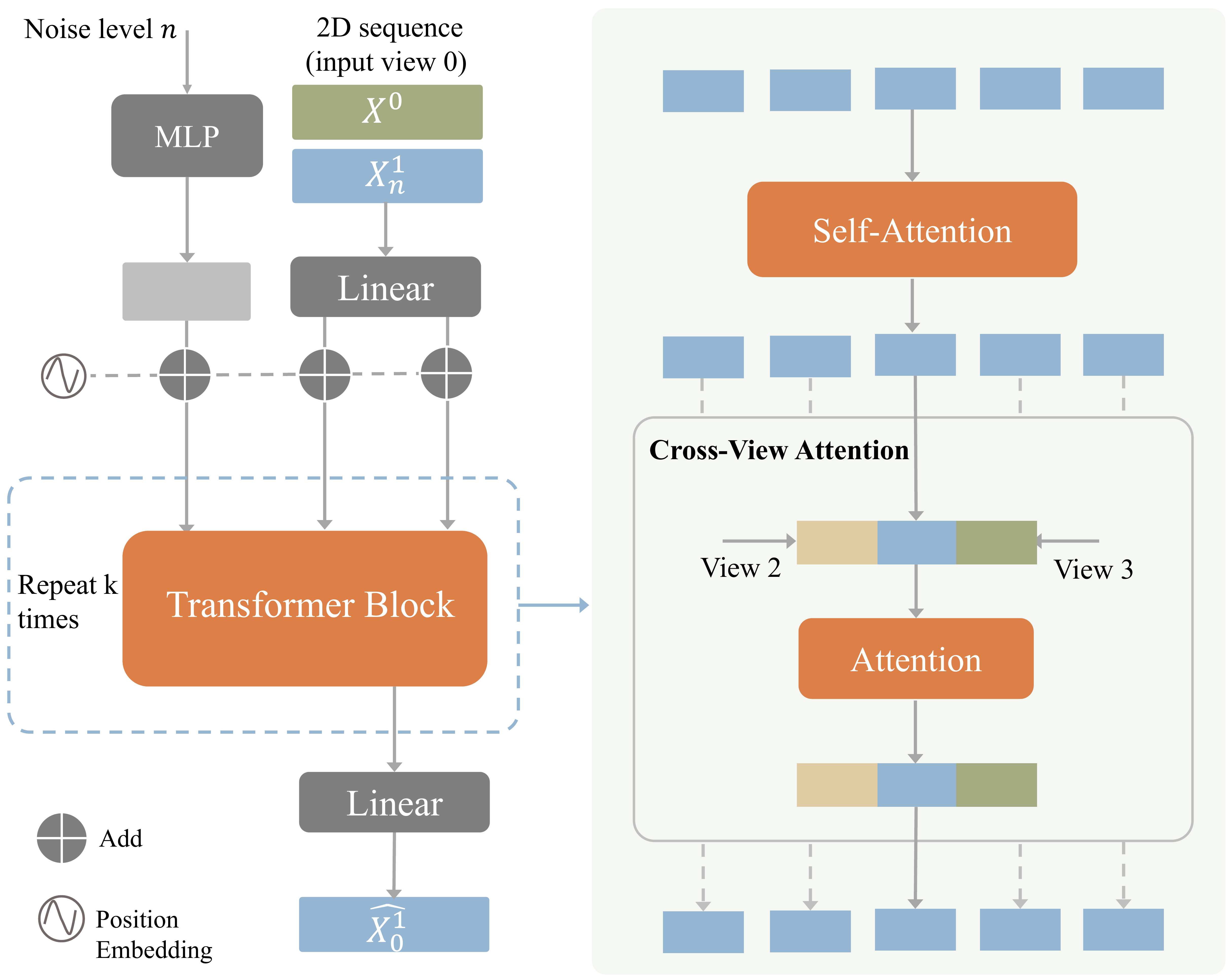}
\end{center}
\vspace{-6mm}
  \caption{Denoising network of the multi-view 2D motion diffusion model, using View 1 as an example for illustration.}
  \label{fig:mv_attn}
\vspace{-3mm}
\end{figure}

\myparagraph{Model Architecture.}
We extend the conditional diffusion model from Stage 1 to handle multi-view generation. Given an input 2D pose sequence $\bm{X}^0$ of view 0, our model generates corresponding sequences for three additional views, denoted as $\bm{X}^{k}, k = 1, 2, 3$. The denoising network uses a transformer-based architecture where each transformer block is augmented with a cross-view attention layer following its self-attention layer, as illustrated in Figure~\ref{fig:mv_attn}.



\begin{table*}[t!]
\small
\begin{center}
\setlength{\tabcolsep}{4pt}
\begin{tabular}{lccccccccccccccc} 
 \toprule 
 & \multicolumn{6}{c}{AIST++~\cite{li2021ai}} & \multicolumn{3}{c}{Steezy} & \multicolumn{3}{c}{NicoleMove} & \multicolumn{3}{c}{CatPlay}  \\
 \cmidrule(lr){2-7}\cmidrule(lr){8-10}\cmidrule(lr){11-13}\cmidrule(lr){14-16} 
 Method    & $\mathbf{T}_{root}$ & MPJPE & PA-MPJPE & $\mathbf{J}_{2D}$ & $\mathbf{J}_{2D}^{C}$ & FID  & $\mathbf{J}_{2D}$ & $\mathbf{J}_{2D}^{C}$ & FID & $\mathbf{J}_{2D}$ & $\mathbf{J}_{2D}^{C}$  & FID & $\mathbf{J}_{2D}$ & $\mathbf{J}_{2D}^{C}$ & FID  \\
        \midrule
        MotionBERT~\cite{zhu2023motionbert} & 101.6 & 134.0 & 108.6 & 59.6 & 32.3 & 3.6 & 67.5 & 31.3 & 14.1 & 78.8 & 62.5 & 6.4 & N/A & N/A & N/A \\
        WHAM~\cite{shin2024wham} & 164.3 & \textbf{104.8} & \textbf{75.1} & 94.3 & 24.1 & 2.6 & 60.8 & 22.5 & \textbf{12.1} & 69.4 & 43.0 & 6.9 & N/A & N/A & N/A \\
        \midrule
        ElePose~\cite{wandt2022elepose} & N/A & 269.4 & 215.1 & 104.9 & 57.1 & 4.8 & 132.7 & 47.6 & 13.3 & 157.4 & 91.1 & 13.3 & N/A & N/A & N/A  \\
        SMPLify~\cite{smplify} & 77.4 & 171.6 & 146.7 & 28.8 & 18.8 & 2.6 & 22.8 & 14.3 & 13.1 & 45.1 & 28.6 & 7.8 & 58.8 & 42.0 & 12.4  \\
        MAS~\cite{kapon2024mas} & N/A & 191.1 & 155.6 & 81.5 & 38.1 & 4.4 & 106.4 & 61.6 & 16.1 & 100.1 & 56.4 & 6.3 & 180.6 & 59.7 & 22.0 \\
        MVLift (ours) & \textbf{67.6} & 110.7 & 79.2 & \textbf{14.0} & \textbf{12.8} & \textbf{1.9} & \textbf{11.7} & \textbf{11.7} & 12.4 & \textbf{26.2} & \textbf{22.7} & \textbf{5.1} & \textbf{57.0} & \textbf{39.4} & \textbf{8.5}  \\
        \bottomrule
\end{tabular}
\end{center}
\vspace{-5mm}
\caption{Results of human pose lifting and animal pose lifting. The baselines MotionBERT~\cite{zhu2023motionbert} and WHAM~\cite{shin2024wham} require training on ground truth 3D motions, while the other approaches do not rely on 3D data for training.} 
    \label{tab:human_pose_estimation}
\vspace{-4mm}
\end{table*}

\section{Experiments}
We first introduce the datasets and evaluation metrics used. Then, we compare our approach against prior work across various datasets, including human motion, animal motion, and human-object interaction motion data. Additionally, we conduct ablation studies to evaluate each stage. 

\subsection{Datasets and Evaluation Metrics}
\noindent\textbf{AIST++}~\cite{li2021ai} consists of multi-view RGB videos and corresponding 3D dance motions. We use this dataset to evaluate the accuracy of predicted 3D joint positions. During training, we randomly select one video from each sequence and extract 2D poses using ViTPose~\cite{xu2022vitpose}, intentionally excluding videos from other camera views to maintain a setting where only single-view videos are available. 

\noindent\textbf{Steezy} consists of dance videos from an online dance teaching website and was originally collected by prior work to study dance motion synthesis~\cite{li2020learning}. We processed the videos using ViTPose~\cite{xu2022vitpose} to obtain 2D pose sequences.

\noindent\textbf{NicoleMove} is our newly collected dataset, which consists of in-the-wild videos from a YouTube channel. We extracted 2D motion sequences using ViTPose~\cite{xu2022vitpose}. The dataset primarily contains motions for yoga, pilates, and fitness.

\noindent\textbf{CatPlay} is also our newly collected dataset, which contains videos of cats playing, captured in an indoor environment. We processed these videos using an advanced approach for generic 2D keypoint estimation~\cite{yang2023unipose}.

\noindent\textbf{OMOMO}~\cite{li2023object} consists of synchronized 3D human motions and object motions. We evaluate our approach on a subset of OMOMO that includes interactions with a specific object, the \textit{largebox}. As the dataset does not include RGB videos, for each sequence, we randomly reproject the 3D motion into a single-view 2D sequence to form a dataset containing 2D human poses and 2D object keypoints.

\myparagraph{Evaluation Metrics.}
We conduct evaluations on datasets with available 3D  motion (AIST++, OMOMO) and datasets with only 2D motion (Steezy, NicoleMove, CatPlay). 
For datasets with 3D motion, we evaluate the predicted 3D motion by computing the root translation error ($\mathbf{T}_{\text{root}}$), the mean per-joint position error (\textbf{MPJPE}), and the mean per-joint position error with Procrustes Alignment (\textbf{PA-MPJPE}), following standard pose estimation protocols~\cite{shin2024wham}. For datasets with only 2D motion, we assess the reprojected 2D motion from the predicted 3D. That is, we calculate the 2D joint position error ($\mathbf{J}_{2D}$) and the 2D joint position error with the 2D root joint translated to the image center ($\mathbf{J}_{2D}^{C}$). Additionally, we train a 2D motion feature extractor on each dataset and use it to compute the Fréchet Inception Distance (\textbf{FID}) for 2D motions reprojected in other camera views, excluding those from the input view. For the object interaction dataset, OMOMO~\cite{li2023object}, we include two additional metrics: $\mathbf{T}_{\text{root}}^{\text{O}}$, which denotes the object translation error, and $\textbf{O-MPJPE}$, which represents the error in 3D object keypoints with the object translated to the origin.

\subsection{Human Pose Lifting}
\paragraph{Baselines.}
We compare our approach with two categories of baselines. The first category includes baselines that require training their model on 3D ground truth motion data, which provides them with a significant data advantage. This category includes MotionBERT~\cite{zhu2023motionbert}, which is trained solely on AMASS, and WHAM~\cite{shin2024wham}, which utilizes AMASS~\cite{AMASS} along with various other datasets containing paired videos and 3D human motions. The second category of baselines does not rely on 3D motion data for training. SMPLify~\cite{smplify} conducts optimization using 2D reprojection objectives to obtain SMPL~\cite{smpl} parameters. ElePose~\cite{wandt2022elepose}, designed for single 3D pose lifting, proposes learning a 2D-to-3D lifting network using reprojection loss and a 2D pose prior loss. MAS~\cite{kapon2024mas} leverages a 2D motion diffusion model to perform unconditional 3D motion generation. We modified MAS to take an input 2D pose sequence as an optimization objective for comparison. We train ElePose and MAS for each dataset using their official codebases.

\begin{figure*}[t!]
\begin{center}
\includegraphics[width=\textwidth]{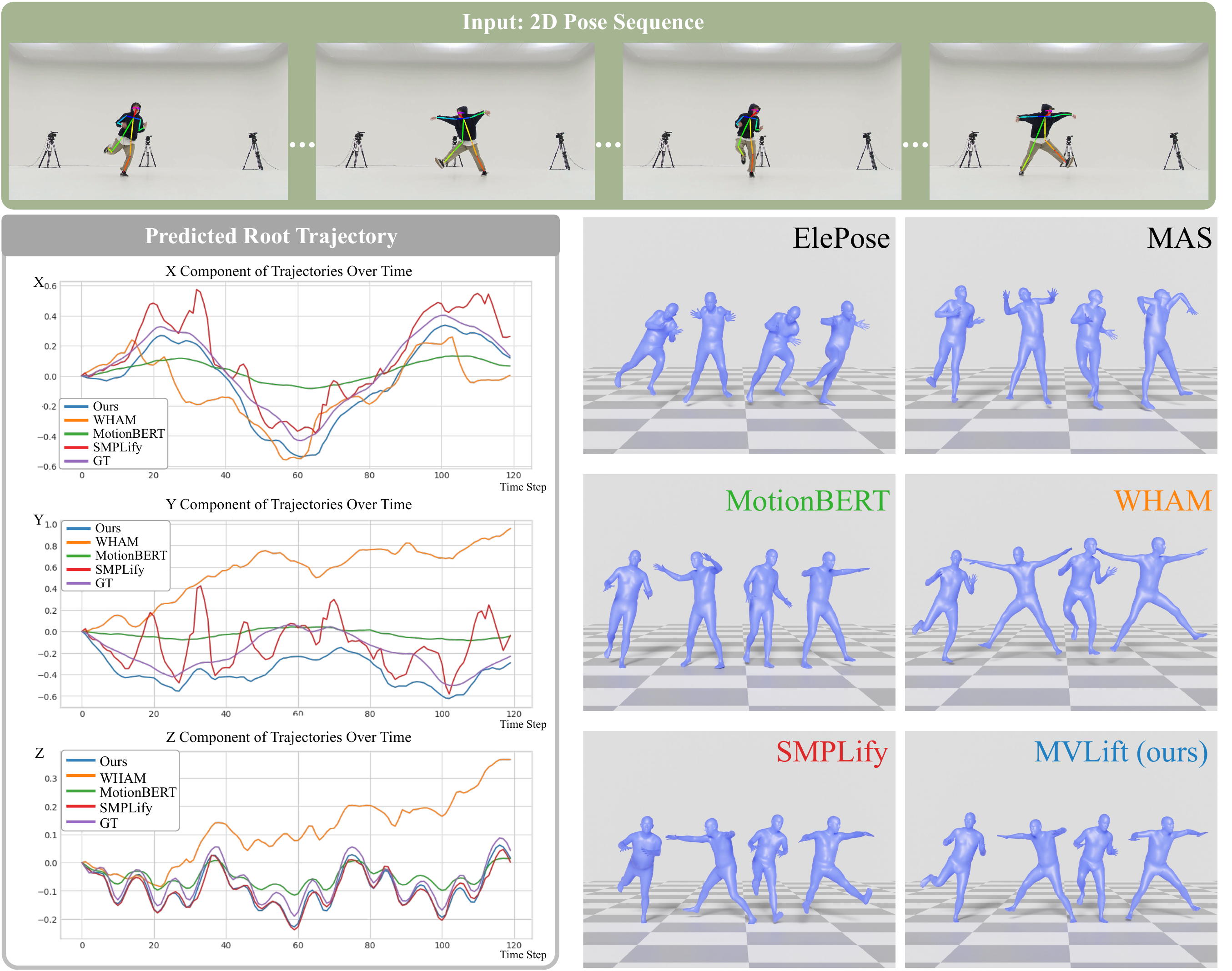}
\end{center}
\vspace{-6mm}
\caption{Qualitative results of AIST++. On the left, we show the trajectories for each component $(x, y, z)$ of the predicted root trajectory. 
}
\label{fig:res_cmp}
\vspace{-4mm}
\end{figure*}

\myparagraph{Results.}
We evaluate our approach and baselines on the test sets of AIST++, Steezy and NicoleMove, as shown in Table~\ref{tab:human_pose_estimation}. We outperform all baselines across several metrics, including those evaluating root trajectory and reprojected 2D poses of the input and other camera views. In terms of 3D joint position errors, our results are superior to all methods that do not require 3D motion data during training, and we also outperform MotionBERT~\cite{zhu2023motionbert}, which requires training on AMASS~\cite{AMASS}. We achieve comparable results to the baseline WHAM~\cite{shin2024wham} in 3D joint position errors with a significant improvement in the root translation metric. 

\begin{figure}[t!]
\begin{center}
\includegraphics[width=\linewidth]{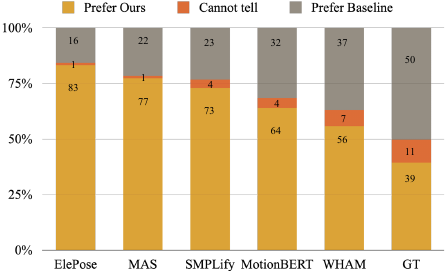}
\end{center}
\vspace{-6mm}
  \caption{Results of human perceptual studies.}
  \label{fig:human_study}
\vspace{-5mm}
\end{figure}

We showcase qualitative comparisons for AIST++ in Figure~\ref{fig:res_cmp}. Note that ElePose and MAS cannot predict root trajectory. ElePose often generates unrealistic 3D poses due to training instability because its pre-trained 2D pose prior is insufficient to prevent unrealistic 2D poses. MAS predicts mean poses that poorly match the input 2D poses because its unconditional 2D motion diffusion model struggles to produce consistent multi-view 2D sequences essential for accurate 3D motion optimization. MotionBERT fails to produce highly dynamic 3D motions as it is trained on the AMASS dataset, which limits its performance on motions outside the dataset distribution. WHAM produces unrealistic root trajectories, deviating from ground truth, because it is trained on constrained 3D motion datasets and cannot generalize well beyond this data. SMPLify, optimizing only based on 2D joint positions, often results in 3D poses with sudden and unrealistic depth changes due to unresolved depth ambiguity.  

\myparagraph{Human Perceptual Study.}
We conducted a human perceptual study to complement the evaluation of generated 3D motion quality. For each dataset (AIST++, Steezy, NicoleMove), we selected 15 generated motion sequences from each approach and created pairs comparing our approach with each baseline. The evaluation was conducted using Amazon Mechanical Turk (AMT). As shown in Figure~\ref{fig:human_study}, participants preferred our results over all baselines. Additionally, when comparing our AIST++ results with ground truth motions, 39\% of participants preferred our results, and 11\% found no noticeable difference between ours and ground truth, demonstrating the realism of our generated sequences.

\subsection{Animal Pose Lifting}
\paragraph{Baselines.}
Since some baselines used in evaluating human pose lifting are designed specifically for human pose predictions, we compare our approach with two baselines that are suitable for animal poses. These include MAS~\cite{kapon2024mas} and modified SMPLify~\cite{smplify}, adapted to use the SMAL model~\cite{Zuffi:CVPR:2017}.

\begin{figure}[t!]
\begin{center}
\includegraphics[width=\linewidth]{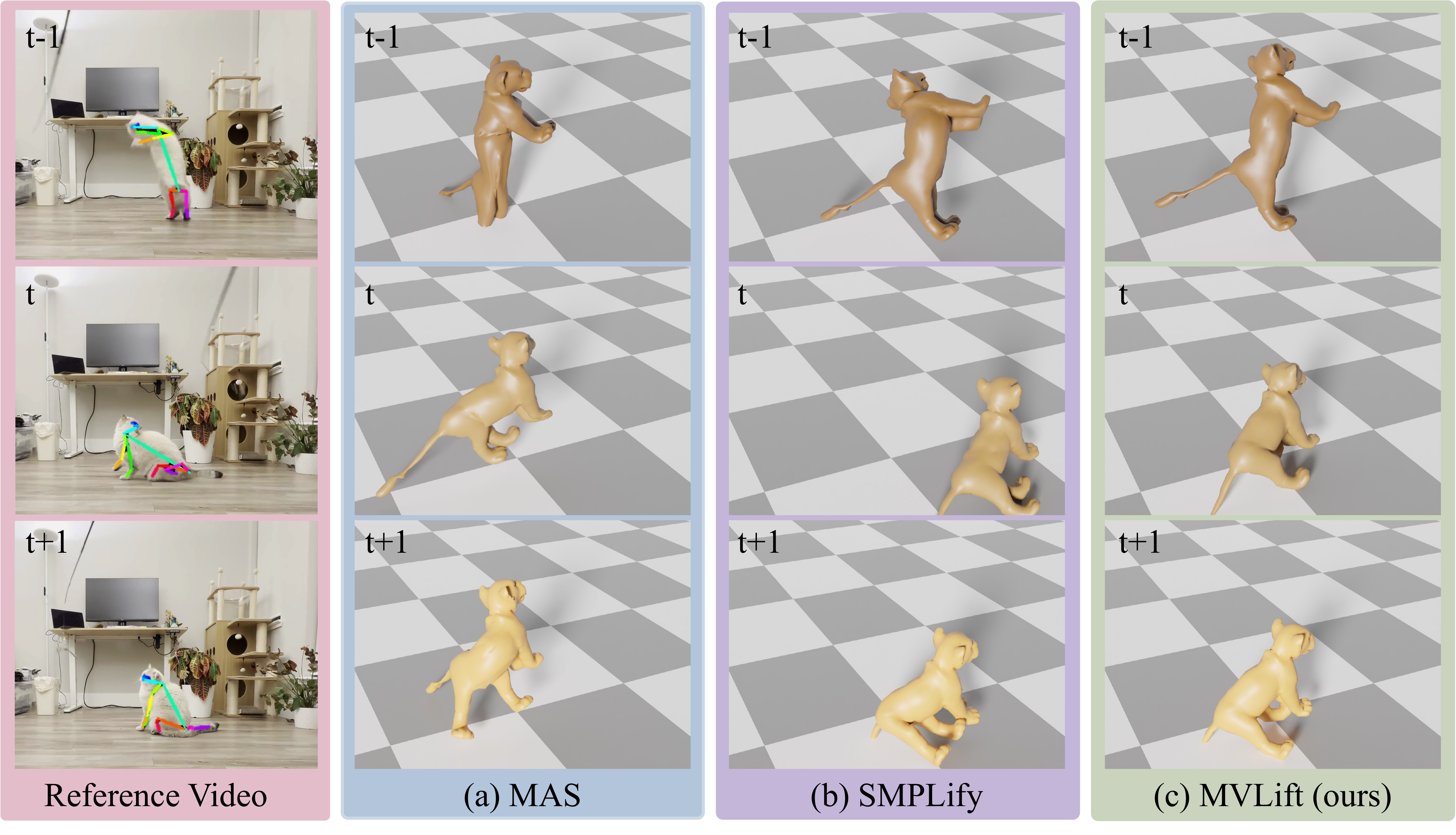}
\end{center}
\vspace{-6mm}
  \caption{Qualitative result comparisons of CatPlay. }
  \label{fig:cat_cmp}
\end{figure}


\myparagraph{Results.}
As shown in Table~\ref{tab:human_pose_estimation}, we outperform all baselines. Qualitative results are showcased in Figure~\ref{fig:cat_cmp}. While MAS cannot generate 3D motions with root trajectories and SMPLify produces unrealistic depth changes due to its single-view limitation, our method generates realistic 3D motions with natural root trajectories.

\subsection{Human-Object Interaction Lifting}

\paragraph{Baselines.} Since no existing work has addressed the task of predicting 3D human-object interaction motions in a world coordinate system from 2D object keypoints and 2D human poses, we adapt SMPLify as a baseline for comparison. This baseline reconstructs 3D human motion and object motion by optimizing SMPL parameters and 3D object pose to match the input 2D keypoint positions of human and object.

\begin{figure}[t!]
\begin{center}
\includegraphics[width=\linewidth]{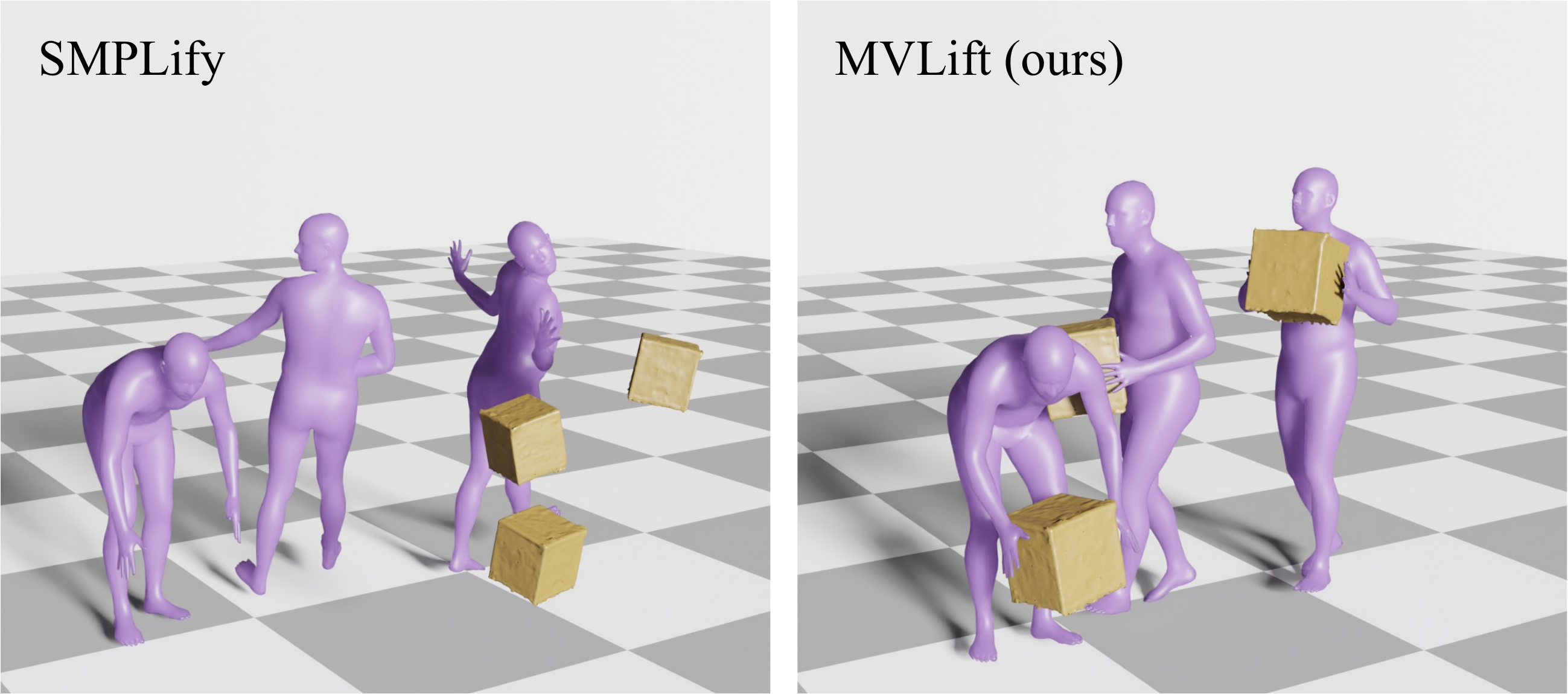}
\end{center}
\vspace{-4mm}
  \caption{Qualitative result comparisons of OMOMO.}
  \label{fig:omomo_cmp}
\end{figure}

\myparagraph{Results.}
We present comparisons in Table~\ref{tab:hoi} and Figure~\ref{fig:omomo_cmp}. SMPLify fails to produce plausible human-object interactions due to significant depth ambiguity from single-view observations. Its predictions lack realistic hand-object contact, and the predicted object motions often contain abrupt changes along the depth direction. Our approach produces realistic human-object interactions by leveraging our generated multi-view 2D pose sequences, which are geometrically consistent across views, for 3D interaction optimization.

\begin{table}[t!]
\small
\begin{center}
\footnotesize{
\setlength{\tabcolsep}{4pt}
\begin{tabular}{lcccccc} 
 \toprule 
 Method    & $\mathbf{T}_{root}$ & MPJPE & PA-MPJPE & $\mathbf{T}_{root}^{O}$ & O-MPJPE     \\
        \midrule
        SMPLify & 97.9 & 142.0 & 99.3 & 751.8 & 106.7   \\
        MVLift (ours) & \textbf{54.9} & \textbf{67.0} & \textbf{53.4} & \textbf{172.9} & \textbf{76.9}  \\
        \bottomrule
\end{tabular}
}
\end{center}
\vspace{-5mm}
\caption{Results of interaction lifting on OMOMO~\cite{li2023object}.} 
    \label{tab:hoi}
\end{table}

\subsection{Ablation Study}
We conduct ablations on the AIST++ dataset, which contains 3D ground truth motions for comprehensive evaluation. We present comparisons in Table~\ref{tab:ablation_study}.

\myparagraph{Evaluation of Each Stage.}
We compare 3D motions obtained using generated multi-view 2D sequences from different stages. MVLift-Stage 1 uses 2D sequences directly generated from the line-conditioned 2D motion diffusion model using epipolar line conditions. MVLift-Stage 2 uses multi-view 2D sequences optimized with SDS and multi-view consistency loss. As shown in Table~\ref{tab:ablation_study}, our final approach (MVLift) using the learned multi-view 2D motion diffusion model achieves more accurate 3D motion estimation than MVLift-Stage 1 and MVLift-Stage 2.

\myparagraph{Evaluation of Stage 2 Ablations.}
In Stage 2, we optimize multi-view 2D pose sequences using SDS and consistency loss. One alternative approach is directly optimizing 3D motion for multi-view consistency while using SDS on the reprojected multi-view 2D sequences to maintain motion realism (SDS for 3D). However, this approach causes the reprojected 2D poses in the input camera view to deviate from the input 2D sequence, resulting in inaccurate 3D motions. Additionally, we experimented with using SDS loss based on an unconditional 2D motion diffusion model (SDS for 3D, without $l_{\text{epi}}$). This variant failed to generate plausible 3D motions and resulted in large joint position errors.

\begin{table}[t!]
\small
\begin{center}
\footnotesize{
\setlength{\tabcolsep}{6pt}
\begin{tabular}{lcccccc} 
 \toprule 
 Method    & $\mathbf{T}_{root}$ & MPJPE & PA-MPJPE & $\mathbf{J}_{2D}$     \\
        \midrule
        \model-Stage 1 & 73.1 & 135.2 & 104.4 & 31.0  \\ 
        \model-Stage 2 & \textbf{65.3} & 127.4 & 96.2 & 19.7  \\ 
        SDS for 3D & 72.9 & 137.3 & 103.5 & 25.2  \\
        SDS for 3D, w/o $l_{epi}$ & 752.3 & 230.4 & 186.2 & 54.9  \\
        \model (ours) & 67.6 & \textbf{110.7} & \textbf{79.2} & \textbf{14.0}  \\
        \bottomrule
\end{tabular}
}
\end{center}
\vspace{-5mm}
\caption{Ablation study on AIST++~\cite{li2021ai}.} 
    \label{tab:ablation_study}
\vspace{-4mm}
\end{table}
\section{Conclusion}
We addressed the challenging problem of estimating 3D motion in the world coordinate system from a 2D motion sequence input, without requiring 3D supervision. The key idea is to combine 2D motion diffusion priors and geometric constraints in a novel multi-stage framework to progressively establish consistency for 2D pose sequences across views. We first trained a line-conditioned 2D motion diffusion model to optimize roughly consistent multi-view 2D sequences, which were then used to create a synthetic dataset with strictly consistent multi-view sequences through 3D optimization and reprojection. Using this dataset, we trained a final diffusion model for efficient multi-view generation. We demonstrated our approach's effectiveness in diverse domains, including human poses, human-object interactions, and animal poses.

\paragraph{Acknowledgement.}
We thank Zhengfei Kuang for fruitful discussions and Waffle and Mochi for CatPlay data collection. This work is in part supported by the Wu Tsai Human Performance Alliance at Stanford University and ONR MURI N00014-22-1-2740.  

{
    \small
    \bibliographystyle{ieeenat_fullname}
    \bibliography{main}

\begin{thebibliography}{67}
\providecommand{\natexlab}[1]{#1}
\providecommand{\url}[1]{\texttt{#1}}
\expandafter\ifx\csname urlstyle\endcsname\relax
  \providecommand{\doi}[1]{doi: #1}\else
  \providecommand{\doi}{doi: \begingroup \urlstyle{rm}\Url}\fi

\bibitem[Araujo et~al.(2023)Araujo, Li, Vetrivel, Agarwal, Gopinath, Wu, Clegg, and Liu]{araujo2023circle}
Joao~Pedro Araujo, Jiaman Li, Karthik Vetrivel, Rishi Agarwal, Deepak Gopinath, Jiajun Wu, Alexander Clegg, and C~Karen Liu.
\newblock {CIRCLE}: Capture in rich contextual environments.
\newblock In \emph{CVPR}, 2023.

\bibitem[Bogo et~al.(2016)Bogo, Kanazawa, Lassner, Gehler, Romero, and Black]{smplify}
Federica Bogo, Angjoo Kanazawa, Christoph Lassner, Peter Gehler, Javier Romero, and Michael~J. Black.
\newblock Keep it {SMPL}: Automatic estimation of {3D} human pose and shape from a single image.
\newblock In \emph{ECCV}, 2016.

\bibitem[Cai et~al.(2019)Cai, Ge, Liu, Cai, Cham, Yuan, and Thalmann]{cai2019exploiting}
Yujun Cai, Liuhao Ge, Jun Liu, Jianfei Cai, Tat-Jen Cham, Junsong Yuan, and Nadia~Magnenat Thalmann.
\newblock Exploiting spatial-temporal relationships for 3d pose estimation via graph convolutional networks.
\newblock In \emph{ICCV}, 2019.

\bibitem[Chan et~al.(2023)Chan, Nagano, Chan, Bergman, Park, Levy, Aittala, De~Mello, Karras, and Wetzstein]{chan2023generative}
Eric~R Chan, Koki Nagano, Matthew~A Chan, Alexander~W Bergman, Jeong~Joon Park, Axel Levy, Miika Aittala, Shalini De~Mello, Tero Karras, and Gordon Wetzstein.
\newblock Generative novel view synthesis with 3d-aware diffusion models.
\newblock In \emph{ICCV}, 2023.

\bibitem[Choi et~al.(2021)Choi, Moon, Chang, and Lee]{choi2021beyond}
Hongsuk Choi, Gyeongsik Moon, Ju~Yong Chang, and Kyoung~Mu Lee.
\newblock Beyond static features for temporally consistent 3d human pose and shape from a video.
\newblock In \emph{CVPR}, 2021.

\bibitem[Ci et~al.(2019)Ci, Wang, Ma, and Wang]{ci2019optimizing}
Hai Ci, Chunyu Wang, Xiaoxuan Ma, and Yizhou Wang.
\newblock Optimizing network structure for 3d human pose estimation.
\newblock In \emph{ICCV}, 2019.

\bibitem[Deitke et~al.(2023)Deitke, Schwenk, Salvador, Weihs, Michel, VanderBilt, Schmidt, Ehsani, Kembhavi, and Farhadi]{deitke2023objaverse}
Matt Deitke, Dustin Schwenk, Jordi Salvador, Luca Weihs, Oscar Michel, Eli VanderBilt, Ludwig Schmidt, Kiana Ehsani, Aniruddha Kembhavi, and Ali Farhadi.
\newblock Objaverse: A universe of annotated 3d objects.
\newblock In \emph{CVPR}, 2023.

\bibitem[Goel et~al.(2023)Goel, Pavlakos, Rajasegaran, Kanazawa, and Malik]{goel2023humans}
Shubham Goel, Georgios Pavlakos, Jathushan Rajasegaran, Angjoo Kanazawa, and Jitendra Malik.
\newblock Humans in 4d: Reconstructing and tracking humans with transformers.
\newblock In \emph{ICCV}, 2023.

\bibitem[Ho et~al.(2020)Ho, Jain, and Abbeel]{ho2020denoising}
Jonathan Ho, Ajay Jain, and Pieter Abbeel.
\newblock Denoising diffusion probabilistic models.
\newblock In \emph{NeurIPS}, 2020.

\bibitem[Ionescu et~al.(2013)Ionescu, Papava, Olaru, and Sminchisescu]{ionescu2013human3}
Catalin Ionescu, Dragos Papava, Vlad Olaru, and Cristian Sminchisescu.
\newblock Human3. 6m: Large scale datasets and predictive methods for 3d human sensing in natural environments.
\newblock \emph{IEEE transactions on pattern analysis and machine intelligence}, 36\penalty0 (7):\penalty0 1325--1339, 2013.

\bibitem[Jiang et~al.(2024)Jiang, Zhang, Li, Ma, Wang, Chen, Liu, Zhu, and Huang]{jiang2024scaling}
Nan Jiang, Zhiyuan Zhang, Hongjie Li, Xiaoxuan Ma, Zan Wang, Yixin Chen, Tengyu Liu, Yixin Zhu, and Siyuan Huang.
\newblock Scaling up dynamic human-scene interaction modeling.
\newblock In \emph{CVPR}, 2024.

\bibitem[Kanazawa et~al.(2018)Kanazawa, Black, Jacobs, and Malik]{kanazawa2018end}
Angjoo Kanazawa, Michael~J Black, David~W Jacobs, and Jitendra Malik.
\newblock End-to-end recovery of human shape and pose.
\newblock In \emph{CVPR}, 2018.

\bibitem[Kanazawa et~al.(2019)Kanazawa, Zhang, Felsen, and Malik]{kanazawa2019learning}
Angjoo Kanazawa, Jason~Y Zhang, Panna Felsen, and Jitendra Malik.
\newblock Learning 3d human dynamics from video.
\newblock In \emph{CVPR}, 2019.

\bibitem[Kapon et~al.(2024)Kapon, Tevet, Cohen-Or, and Bermano]{kapon2024mas}
Roy Kapon, Guy Tevet, Daniel Cohen-Or, and Amit~H Bermano.
\newblock Mas: Multi-view ancestral sampling for 3d motion generation using 2d diffusion.
\newblock In \emph{CVPR}, 2024.

\bibitem[Kocabas et~al.(2020)Kocabas, Athanasiou, and Black]{kocabas2019vibe}
Muhammed Kocabas, Nikos Athanasiou, and Michael~J. Black.
\newblock Vibe: Video inference for human body pose and shape estimation.
\newblock In \emph{CVPR}, 2020.

\bibitem[Kocabas et~al.(2021{\natexlab{a}})Kocabas, Huang, Hilliges, and Black]{kocabas2021pare}
Muhammed Kocabas, Chun-Hao~P Huang, Otmar Hilliges, and Michael~J Black.
\newblock Pare: Part attention regressor for 3d human body estimation.
\newblock In \emph{ICCV}, 2021{\natexlab{a}}.

\bibitem[Kocabas et~al.(2021{\natexlab{b}})Kocabas, Huang, Tesch, M{\"u}ller, Hilliges, and Black]{kocabas2021spec}
Muhammed Kocabas, Chun-Hao~P Huang, Joachim Tesch, Lea M{\"u}ller, Otmar Hilliges, and Michael~J Black.
\newblock Spec: Seeing people in the wild with an estimated camera.
\newblock In \emph{ICCV}, 2021{\natexlab{b}}.

\bibitem[Kocabas et~al.(2024)Kocabas, Yuan, Molchanov, Guo, Black, Hilliges, Kautz, and Iqbal]{kocabas2024pace}
Muhammed Kocabas, Ye Yuan, Pavlo Molchanov, Yunrong Guo, Michael~J Black, Otmar Hilliges, Jan Kautz, and Umar Iqbal.
\newblock Pace: Human and motion estimation from in-the-wild videos.
\newblock \emph{3DV}, 2024.

\bibitem[Kolotouros et~al.(2019)Kolotouros, Pavlakos, Black, and Daniilidis]{kolotouros2019learning}
Nikos Kolotouros, Georgios Pavlakos, Michael~J Black, and Kostas Daniilidis.
\newblock Learning to reconstruct 3d human pose and shape via model-fitting in the loop.
\newblock In \emph{ICCV}, 2019.

\bibitem[Kuang et~al.(2024)Kuang, Cai, He, Xu, Li, Guibas, and Wetzstein]{kuang2024collaborative}
Zhengfei Kuang, Shengqu Cai, Hao He, Yinghao Xu, Hongsheng Li, Leonidas Guibas, and Gordon Wetzstein.
\newblock Collaborative video diffusion: Consistent multi-video generation with camera control.
\newblock \emph{arXiv preprint arXiv:2405.17414}, 2024.

\bibitem[Li et~al.(2020)Li, Yin, Chu, Zhou, Wang, Fidler, and Li]{li2020learning}
Jiaman Li, Yihang Yin, Hang Chu, Yi Zhou, Tingwu Wang, Sanja Fidler, and Hao Li.
\newblock Learning to generate diverse dance motions with transformer.
\newblock \emph{arXiv preprint arXiv:2008.08171}, 2020.

\bibitem[Li et~al.(2021{\natexlab{a}})Li, Xu, Chen, Bian, Yang, and Lu]{li2021hybrik}
Jiefeng Li, Chao Xu, Zhicun Chen, Siyuan Bian, Lixin Yang, and Cewu Lu.
\newblock Hybrik: A hybrid analytical-neural inverse kinematics solution for 3d human pose and shape estimation.
\newblock In \emph{CVPR}, 2021{\natexlab{a}}.

\bibitem[Li et~al.(2023{\natexlab{a}})Li, Bian, Liu, Tang, Wang, and Lu]{li2023niki}
Jiefeng Li, Siyuan Bian, Qi Liu, Jiasheng Tang, Fan Wang, and Cewu Lu.
\newblock Niki: Neural inverse kinematics with invertible neural networks for 3d human pose and shape estimation.
\newblock In \emph{CVPR}, 2023{\natexlab{a}}.

\bibitem[Li et~al.(2023{\natexlab{b}})Li, Wu, and Liu]{li2023object}
Jiaman Li, Jiajun Wu, and C~Karen Liu.
\newblock Object motion guided human motion synthesis.
\newblock \emph{ACM Trans. Graph.}, 42\penalty0 (6), 2023{\natexlab{b}}.

\bibitem[Li et~al.(2021{\natexlab{b}})Li, Yang, Ross, and Kanazawa]{li2021ai}
Ruilong Li, Shan Yang, David~A Ross, and Angjoo Kanazawa.
\newblock Ai choreographer: Music conditioned 3d dance generation with aist++.
\newblock In \emph{ICCV}, 2021{\natexlab{b}}.

\bibitem[Li et~al.(2022)Li, Liu, Tang, Wang, and Van~Gool]{li2022mhformer}
Wenhao Li, Hong Liu, Hao Tang, Pichao Wang, and Luc Van~Gool.
\newblock Mhformer: Multi-hypothesis transformer for 3d human pose estimation.
\newblock In \emph{CVPR}, 2022.

\bibitem[Liu et~al.(2023{\natexlab{a}})Liu, Wu, Van~Hoorick, Tokmakov, Zakharov, and Vondrick]{liu2023zero}
Ruoshi Liu, Rundi Wu, Basile Van~Hoorick, Pavel Tokmakov, Sergey Zakharov, and Carl Vondrick.
\newblock Zero-1-to-3: Zero-shot one image to 3d object.
\newblock In \emph{ICCV}, 2023{\natexlab{a}}.

\bibitem[Liu et~al.(2023{\natexlab{b}})Liu, Lin, Zeng, Long, Liu, Komura, and Wang]{liu2023syncdreamer}
Yuan Liu, Cheng Lin, Zijiao Zeng, Xiaoxiao Long, Lingjie Liu, Taku Komura, and Wenping Wang.
\newblock Syncdreamer: Generating multiview-consistent images from a single-view image.
\newblock \emph{arXiv preprint arXiv:2309.03453}, 2023{\natexlab{b}}.

\bibitem[Long et~al.(2024)Long, Guo, Lin, Liu, Dou, Liu, Ma, Zhang, Habermann, Theobalt, et~al.]{long2024wonder3d}
Xiaoxiao Long, Yuan-Chen Guo, Cheng Lin, Yuan Liu, Zhiyang Dou, Lingjie Liu, Yuexin Ma, Song-Hai Zhang, Marc Habermann, Christian Theobalt, et~al.
\newblock Wonder3d: Single image to 3d using cross-domain diffusion.
\newblock In \emph{CVPR}, 2024.

\bibitem[Loper et~al.(2015)Loper, Mahmood, Romero, Pons-Moll, and Black]{smpl}
Matthew Loper, Naureen Mahmood, Javier Romero, Gerard Pons-Moll, and Michael~J Black.
\newblock Smpl: A skinned multi-person linear model.
\newblock \emph{ACM transactions on graphics (TOG)}, 34\penalty0 (6):\penalty0 1--16, 2015.

\bibitem[Luo et~al.(2020)Luo, Golestaneh, and Kitani]{luo20203d}
Zhengyi Luo, S~Alireza Golestaneh, and Kris~M Kitani.
\newblock 3d human motion estimation via motion compression and refinement.
\newblock In \emph{ACCV}, 2020.

\bibitem[Mahmood et~al.(2019)Mahmood, Ghorbani, Troje, Pons-Moll, and Black]{AMASS}
Naureen Mahmood, Nima Ghorbani, Nikolaus~F Troje, Gerard Pons-Moll, and Michael~J Black.
\newblock {AMASS}: Archive of motion capture as surface shapes.
\newblock In \emph{ICCV}, 2019.

\bibitem[Martinez et~al.(2017)Martinez, Hossain, Romero, and Little]{martinez2017simple}
Julieta Martinez, Rayat Hossain, Javier Romero, and James~J Little.
\newblock A simple yet effective baseline for 3d human pose estimation.
\newblock In \emph{ICCV}, 2017.

\bibitem[Mehta et~al.(2017)Mehta, Rhodin, Casas, Fua, Sotnychenko, Xu, and Theobalt]{mehta2017monocular}
Dushyant Mehta, Helge Rhodin, Dan Casas, Pascal Fua, Oleksandr Sotnychenko, Weipeng Xu, and Christian Theobalt.
\newblock Monocular 3d human pose estimation in the wild using improved cnn supervision.
\newblock In \emph{3DV}, 2017.

\bibitem[Pavlakos et~al.(2019{\natexlab{a}})Pavlakos, Choutas, Ghorbani, Bolkart, Osman, Tzionas, and Black]{SMPL-X:2019}
Georgios Pavlakos, Vasileios Choutas, Nima Ghorbani, Timo Bolkart, Ahmed A.~A. Osman, Dimitrios Tzionas, and Michael~J. Black.
\newblock Expressive body capture: 3d hands, face, and body from a single image.
\newblock In \emph{CVPR}, 2019{\natexlab{a}}.

\bibitem[Pavlakos et~al.(2019{\natexlab{b}})Pavlakos, Choutas, Ghorbani, Bolkart, Osman, Tzionas, and Black]{smplx}
Georgios Pavlakos, Vasileios Choutas, Nima Ghorbani, Timo Bolkart, Ahmed A.~A. Osman, Dimitrios Tzionas, and Michael~J. Black.
\newblock Expressive body capture: 3{D} hands, face, and body from a single image.
\newblock In \emph{CVPR}, 2019{\natexlab{b}}.

\bibitem[Pavllo et~al.(2019)Pavllo, Feichtenhofer, Grangier, and Auli]{pavllo20193d}
Dario Pavllo, Christoph Feichtenhofer, David Grangier, and Michael Auli.
\newblock 3d human pose estimation in video with temporal convolutions and semi-supervised training.
\newblock In \emph{CVPR}, 2019.

\bibitem[Poole et~al.(2022)Poole, Jain, Barron, and Mildenhall]{poole2022dreamfusion}
Ben Poole, Ajay Jain, Jonathan~T Barron, and Ben Mildenhall.
\newblock Dreamfusion: Text-to-3d using 2d diffusion.
\newblock \emph{arXiv preprint arXiv:2209.14988}, 2022.

\bibitem[Shan et~al.(2022)Shan, Liu, Zhang, Wang, Ma, and Gao]{shan2022p}
Wenkang Shan, Zhenhua Liu, Xinfeng Zhang, Shanshe Wang, Siwei Ma, and Wen Gao.
\newblock P-stmo: Pre-trained spatial temporal many-to-one model for 3d human pose estimation.
\newblock In \emph{ECCV}, 2022.

\bibitem[Shen et~al.(2023)Shen, Yang, Wang, Ma, Zhou, and Yang]{shen2023global}
Xiaolong Shen, Zongxin Yang, Xiaohan Wang, Jianxin Ma, Chang Zhou, and Yi Yang.
\newblock Global-to-local modeling for video-based 3d human pose and shape estimation.
\newblock In \emph{CVPR}, 2023.

\bibitem[Shi et~al.(2023{\natexlab{a}})Shi, Chen, Zhang, Liu, Xu, Wei, Chen, Zeng, and Su]{shi2023zero123++}
Ruoxi Shi, Hansheng Chen, Zhuoyang Zhang, Minghua Liu, Chao Xu, Xinyue Wei, Linghao Chen, Chong Zeng, and Hao Su.
\newblock Zero123++: a single image to consistent multi-view diffusion base model.
\newblock \emph{arXiv preprint arXiv:2310.15110}, 2023{\natexlab{a}}.

\bibitem[Shi et~al.(2023{\natexlab{b}})Shi, Wang, Ye, Long, Li, and Yang]{shi2023mvdream}
Yichun Shi, Peng Wang, Jianglong Ye, Mai Long, Kejie Li, and Xiao Yang.
\newblock Mvdream: Multi-view diffusion for 3d generation.
\newblock \emph{arXiv preprint arXiv:2308.16512}, 2023{\natexlab{b}}.

\bibitem[Shin et~al.(2024)Shin, Kim, Halilaj, and Black]{shin2024wham}
Soyong Shin, Juyong Kim, Eni Halilaj, and Michael~J Black.
\newblock Wham: Reconstructing world-grounded humans with accurate 3d motion.
\newblock In \emph{CVPR}, 2024.

\bibitem[Sigal et~al.(2010)Sigal, Balan, and Black]{sigal2010humaneva}
Leonid Sigal, Alexandru~O Balan, and Michael~J Black.
\newblock Humaneva: Synchronized video and motion capture dataset and baseline algorithm for evaluation of articulated human motion.
\newblock \emph{International journal of computer vision}, 87\penalty0 (1):\penalty0 4--27, 2010.

\bibitem[Sun et~al.(2023)Sun, Bao, Liu, Mei, and Black]{sun2023trace}
Yu Sun, Qian Bao, Wu Liu, Tao Mei, and Michael~J Black.
\newblock Trace: 5d temporal regression of avatars with dynamic cameras in 3d environments.
\newblock In \emph{CVPR}, 2023.

\bibitem[Taheri et~al.(2020)Taheri, Ghorbani, Black, and Tzionas]{GRAB:2020}
Omid Taheri, Nima Ghorbani, Michael~J. Black, and Dimitrios Tzionas.
\newblock {GRAB}: A dataset of whole-body human grasping of objects.
\newblock In \emph{ECCV}, 2020.

\bibitem[Tang et~al.(2023)Tang, Wang, Zhang, Zhang, Yi, Ma, and Chen]{tang2023make}
Junshu Tang, Tengfei Wang, Bo Zhang, Ting Zhang, Ran Yi, Lizhuang Ma, and Dong Chen.
\newblock Make-it-3d: High-fidelity 3d creation from a single image with diffusion prior.
\newblock In \emph{ICCV}, 2023.

\bibitem[Tevet et~al.(2023)Tevet, Raab, Gordon, Shafir, Bermano, and Cohen-Or]{tevet2022human}
Guy Tevet, Sigal Raab, Brian Gordon, Yonatan Shafir, Amit~H Bermano, and Daniel Cohen-Or.
\newblock Human motion diffusion model.
\newblock In \emph{ICLR}, 2023.

\bibitem[Tewari et~al.(2023)Tewari, Yin, Cazenavette, Rezchikov, Tenenbaum, Durand, Freeman, and Sitzmann]{tewari2023diffusion}
Ayush Tewari, Tianwei Yin, George Cazenavette, Semon Rezchikov, Josh Tenenbaum, Fr{\'e}do Durand, Bill Freeman, and Vincent Sitzmann.
\newblock Diffusion with forward models: Solving stochastic inverse problems without direct supervision.
\newblock \emph{NeurIPS}, 2023.

\bibitem[Tseng et~al.(2023)Tseng, Castellon, and Liu]{tseng2022edge}
Jonathan Tseng, Rodrigo Castellon, and C~Karen Liu.
\newblock {EDGE}: Editable dance generation from music.
\newblock In \emph{CVPR}, 2023.

\bibitem[Van~Hoorick et~al.(2024)Van~Hoorick, Wu, Ozguroglu, Sargent, Liu, Tokmakov, Dave, Zheng, and Vondrick]{van2024generative}
Basile Van~Hoorick, Rundi Wu, Ege Ozguroglu, Kyle Sargent, Ruoshi Liu, Pavel Tokmakov, Achal Dave, Changxi Zheng, and Carl Vondrick.
\newblock Generative camera dolly: Extreme monocular dynamic novel view synthesis.
\newblock \emph{arXiv preprint arXiv:2405.14868}, 2024.

\bibitem[Vaswani et~al.(2017)Vaswani, Shazeer, Parmar, Uszkoreit, Jones, Gomez, Kaiser, and Polosukhin]{vaswani2017attention}
Ashish Vaswani, Noam Shazeer, Niki Parmar, Jakob Uszkoreit, Llion Jones, Aidan~N Gomez, {\L}ukasz Kaiser, and Illia Polosukhin.
\newblock Attention is all you need.
\newblock In \emph{NIPS}, 2017.

\bibitem[Wan et~al.(2021)Wan, Li, Tian, Liu, Yi, and Li]{wan2021encoder}
Ziniu Wan, Zhengjia Li, Maoqing Tian, Jianbo Liu, Shuai Yi, and Hongsheng Li.
\newblock Encoder-decoder with multi-level attention for 3d human shape and pose estimation.
\newblock In \emph{ICCV}, 2021.

\bibitem[Wandt et~al.(2022)Wandt, Little, and Rhodin]{wandt2022elepose}
Bastian Wandt, James~J Little, and Helge Rhodin.
\newblock Elepose: Unsupervised 3d human pose estimation by predicting camera elevation and learning normalizing flows on 2d poses.
\newblock In \emph{CVPR}, 2022.

\bibitem[Wang et~al.(2020)Wang, Yan, Xiong, and Lin]{wang2020motion}
Jingbo Wang, Sijie Yan, Yuanjun Xiong, and Dahua Lin.
\newblock Motion guided 3d pose estimation from videos.
\newblock In \emph{ECCV}, 2020.

\bibitem[Wei et~al.(2022)Wei, Lin, Liu, and Liao]{wei2022capturing}
Wen-Li Wei, Jen-Chun Lin, Tyng-Luh Liu, and Hong-Yuan~Mark Liao.
\newblock Capturing humans in motion: Temporal-attentive 3d human pose and shape estimation from monocular video.
\newblock In \emph{CVPR}, 2022.

\bibitem[Xu et~al.(2022)Xu, Zhang, Zhang, and Tao]{xu2022vitpose}
Yufei Xu, Jing Zhang, Qiming Zhang, and Dacheng Tao.
\newblock Vitpose: Simple vision transformer baselines for human pose estimation.
\newblock \emph{NeurIPS}, 2022.

\bibitem[Xu et~al.(2023)Xu, Tan, Luan, Bi, Wang, Li, Shi, Sunkavalli, Wetzstein, Xu, et~al.]{xu2023dmv3d}
Yinghao Xu, Hao Tan, Fujun Luan, Sai Bi, Peng Wang, Jiahao Li, Zifan Shi, Kalyan Sunkavalli, Gordon Wetzstein, Zexiang Xu, et~al.
\newblock Dmv3d: Denoising multi-view diffusion using 3d large reconstruction model.
\newblock \emph{arXiv preprint arXiv:2311.09217}, 2023.

\bibitem[Yang et~al.(2023)Yang, Zeng, Zhang, and Zhang]{yang2023unipose}
Jie Yang, Ailing Zeng, Ruimao Zhang, and Lei Zhang.
\newblock Unipose: Detecting any keypoints.
\newblock \emph{arXiv preprint arXiv:2310.08530}, 2023.

\bibitem[Ye et~al.(2023)Ye, Pavlakos, Malik, and Kanazawa]{ye2023decoupling}
Vickie Ye, Georgios Pavlakos, Jitendra Malik, and Angjoo Kanazawa.
\newblock Decoupling human and camera motion from videos in the wild.
\newblock In \emph{CVPR}, 2023.

\bibitem[Yuan et~al.(2022)Yuan, Iqbal, Molchanov, Kitani, and Kautz]{yuan2022glamr}
Ye Yuan, Umar Iqbal, Pavlo Molchanov, Kris Kitani, and Jan Kautz.
\newblock Glamr: Global occlusion-aware human mesh recovery with dynamic cameras.
\newblock In \emph{CVPR}, 2022.

\bibitem[Zhang et~al.(2023{\natexlab{a}})Zhang, Tian, Zhang, Li, An, Sun, and Liu]{zhang2023pymaf}
Hongwen Zhang, Yating Tian, Yuxiang Zhang, Mengcheng Li, Liang An, Zhenan Sun, and Yebin Liu.
\newblock Pymaf-x: Towards well-aligned full-body model regression from monocular images.
\newblock \emph{IEEE Transactions on Pattern Analysis and Machine Intelligence}, 45\penalty0 (10):\penalty0 12287--12303, 2023{\natexlab{a}}.

\bibitem[Zhang et~al.(2022)Zhang, Tu, Yang, Chen, and Yuan]{zhang2022mixste}
Jinlu Zhang, Zhigang Tu, Jianyu Yang, Yujin Chen, and Junsong Yuan.
\newblock Mixste: Seq2seq mixed spatio-temporal encoder for 3d human pose estimation in video.
\newblock In \emph{CVPR}, 2022.

\bibitem[Zhang et~al.(2023{\natexlab{b}})Zhang, Zhang, Hu, Yi, Zhang, and Liu]{zhang2023real}
Yuxiang Zhang, Hongwen Zhang, Liangxiao Hu, Hongwei Yi, Shengping Zhang, and Yebin Liu.
\newblock Real-time monocular full-body capture in world space via sequential proxy-to-motion learning.
\newblock \emph{arXiv preprint arXiv:2307.01200}, 2023{\natexlab{b}}.

\bibitem[Zheng et~al.(2021)Zheng, Zhu, Mendieta, Yang, Chen, and Ding]{zheng20213d}
Ce Zheng, Sijie Zhu, Matias Mendieta, Taojiannan Yang, Chen Chen, and Zhengming Ding.
\newblock 3d human pose estimation with spatial and temporal transformers.
\newblock In \emph{ICCV}, 2021.

\bibitem[Zhu et~al.(2023)Zhu, Ma, Liu, Liu, Wu, and Wang]{zhu2023motionbert}
Wentao Zhu, Xiaoxuan Ma, Zhaoyang Liu, Libin Liu, Wayne Wu, and Yizhou Wang.
\newblock Motionbert: A unified perspective on learning human motion representations.
\newblock In \emph{ICCV}, 2023.

\bibitem[Zuffi et~al.(2017)Zuffi, Kanazawa, Jacobs, and Black]{Zuffi:CVPR:2017}
Silvia Zuffi, Angjoo Kanazawa, David Jacobs, and Michael~J. Black.
\newblock {3D} menagerie: Modeling the {3D} shape and pose of animals.
\newblock In \emph{CVPR}, 2017.

\end{thebibliography}
}


\end{document}